\documentclass[10pt,twocolumn,letterpaper]{article}

\usepackage{wacv}
\usepackage{nopageno}
\usepackage{times}
\usepackage{epsfig}
\usepackage{graphicx}
\usepackage{amsmath}
\usepackage{amssymb}
\usepackage[accsupp]{axessibility}

\def\wacvPaperID{1199} 

\wacvfinalcopy 

\ifwacvfinal

\fi

\ifwacvfinal
\usepackage[breaklinks=true,bookmarks=false]{hyperref}
\else
\usepackage[pagebackref=true,breaklinks=true,colorlinks,bookmarks=false]{hyperref}
\fi

\ifwacvfinal
\else
\fi

\ifwacvfinal\fi
\begin{document}

\title{One-Class Learned Encoder-Decoder Network with Adversarial Context Masking for Novelty Detection}

\author{John Taylor Jewell\\
Western University\\
London, ON, Canada\\
{\tt\small jjewell6@uwo.ca}
\and
Vahid Reza Khazaie\\
Western University\\
London, ON, Canada\\
{\tt\small vkhazaie@uwo.ca}

\and
Yalda Mohsenzadeh\\
Western University\\
London, ON, Canada\\
{\tt\small ymohsenz@uwo.ca}

}

\maketitle

\begin{abstract}
Novelty detection is the task of recognizing samples that do not belong to the distribution of the target class. During training, the novelty class is absent, preventing the use of traditional classification approaches. Deep autoencoders have been widely used as a base of many novelty detection methods. In particular, context autoencoders have been successful in the novelty detection task because of the more effective representations they learn by reconstructing original images from randomly masked images. However, a significant drawback of context autoencoders is that random masking fails to consistently cover important structures of the input image, leading to suboptimal representations - especially for the novelty detection task. In this paper, to optimize input masking, we introduce a Mask Module that learns to generate optimal masks and a Reconstructor that aims to reconstruct masked images. The networks are trained in an adversarial setting in which the Mask Module seeks to maximize the reconstruction error that the Reconstructor is minimizing. When applied to novelty detection, the proposed approach learns semantically richer representations compared to context autoencoders and enhances novelty detection at test time through more optimal masking. Novelty detection experiments on the MNIST and CIFAR-10 image datasets demonstrate the proposed approach's superiority over cutting-edge methods. In a further experiment on the UCSD video dataset for novelty detection, the proposed approach achieves a frame-level Area Under the Curve (AUC) of 99.02\% and an Equal Error Rate (EER) of 5.4\%, exceeding recent state-of-the-art models. Code available at https://github.com/jewelltaylor/OLED.
\end{abstract}

\section{Introduction}

\begin{figure*}[t]
\begin{center}
   \includegraphics[width=1\linewidth]{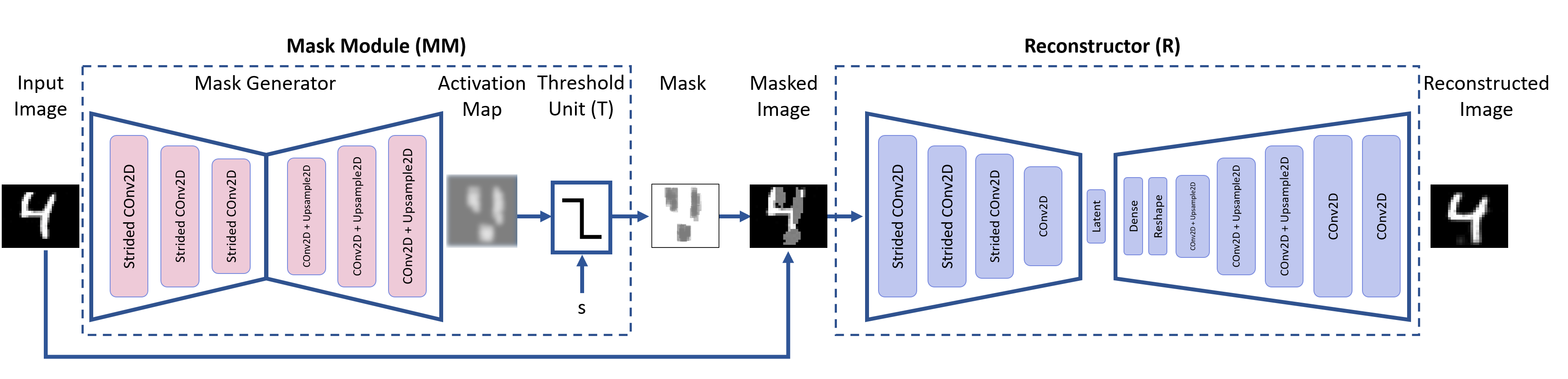}
\end{center}
   \caption{An overview of the architecture in OLED. The Mask Module adversarially learns to cover the important parts of the input image; it consists of an autoencoder that generates an activation map and a threshold unit to produce the binary mask. The Reconstructor aims to minimize the reconstruction error and the Mask Module aims to maximize the reconstruction error.}
\label{model_overview}
\end{figure*}

Novelty detection involves determining whether or not an unknown sample belongs to the distribution of the training data. In the case the sample is similar to the training data, it is referred to as an inlier or normal sample. Alternatively, if the sample does not follow the distribution defined in the training examples, it is referred to as an outlier or anomaly. Novelty detection differs from other machine learning tasks in that the outlier class is poorly sampled or nonexistent. Due to the unavailability of outlier samples, traditional classification approaches are not suitable.

Within computer vision, novelty detection is ubiquitous with subtasks that have widespread applications such as marker discovery in biomedical data \cite{schlegl2017unsupervised} and video surveillance \cite{luo2017revisit}. Anomaly detection in images is one such task that involves identifying whether an image is an inlier or an outlier based on training data that mostly consists of inlier images. To compensate for the unavailability of outlier samples, one-class classification methods aim to model the distribution of the inlier data \cite{zimek2012survey}. New samples that do not conform to the target distribution are considered outliers. However, it is often hard to model the distribution of image data with conventional methods because of the high dimensionality in which the data points reside \cite{zimek2012survey}.

With the advent of deep learning, methods have been proposed that are able to effectively produce representations for high dimensional data \cite{bengio2013representation}. Autoencoders (AE) are an unsupervised class of approaches that are well suited for modeling image data \cite{bengio2007greedy}. At a high level, an AE consists of two modules: an encoder and a decoder. The encoder learns a mapping from an image to a lower-dimensional latent space, and the decoder learns a mapping from the latent space back to the original image. In this way, AEs are trained in an unsupervised manner by minimizing the error between the original image and the reconstruction.

As a powerful unsupervised method for learning representations, AEs are the basis of many one-class classification approaches \cite{chalapathy2019deep}. To detect anomalous images, the AE is first trained on a set of primarily normal images. At test time, the reconstruction error of a sample is used as an anomaly score. The underlying intuition is that the reconstruction error will be lower for inlier samples than outlier samples \cite{xia2015learning}. This follows from the fact that the AE is trained solely on inlier samples. However, this assumption is often violated, and the AE generalizes well to construct images outside of the distribution of the training data \cite{zong2018deep, gong2019memorizing}. This is especially evident in cases where anomalous images share similar compositional patterns as inlier images.

Recent methods introduce additional complexity into the autoencoders reconstruction task so that outliers are not reconstructed well \cite{abati2019latent, perera2019ocgan, zimmerer2018context}. To this end, denoising autoencoders (DAE) have been used. DAEs learn to reconstruct unperturbed images from images that have been perturbed by noise \cite{vincent2010stacked}. Beyond yielding more robust representations, the denoising task of the AE has been shown to induce a reconstruction error that approximates the local derivative of the log-density with respect to the input \cite{alain2014regularized}. Thus, a sample’s global reconstruction error reflects the norm of the derivative of the log-density with respect to the input. In this way, DAEs provide a more interpretable and theoretically grounded anomaly score.

Context autoencoders (CAE) \cite{pathak2016context}, a specific type of DAE, have shown strong performance in the anomaly detection task \cite{zimmerer2018context, baur2020autoencoders}. Instead of being perturbed by noise, input images are subjected to random masking. Consequently, the CAE learns to inpaint a randomly masked region of the input image in conjunction with the reconstruction task. This random masking is similar to using salt-and-pepper noise, which has been shown to yield better representations by implicitly enforcing the AE to learn semantic information about the distribution of the training data \cite{pathak2016context}. Despite these strengths, in some cases CAEs suffer from suboptimal representations leading to poor performance in the anomaly detection task.

Inspired by the drawbacks of CAEs \cite{pathak2016context}, we proposed One-Class Learned Encoder-Decoder (OLED) Network with Adversarial Context Masking. OLED introduces a Mask Module $MM$ that produces masks applied to images input into the Reconstructor $R$. The masks generated by $MM$ are optimized to cover the most important parts of the input image, resulting in a comparable reconstruction score across samples. The underlying intuition is that the loss of the masked region will be low in the case of inlier images and high in the case of outlier images. This stems from the fact that the Reconstructor learns to inpaint masked regions using mostly inlier samples. Thus, the inpainted regions of outlier images will consist of patterns present in the inlier images, yielding a high reconstruction error.

At a high level, the Mask Module is a convolutional autoencoder, and the Reconstructor is a convolutional encoder-decoder. They are trained in an adversarial manner, where the Mask Module is trying to generate masks that yield higher reconstruction errors, and the Reconstructor is trying to minimize the reconstruction error of the masked image. The architecture of the proposed approach is shown in Figure \ref{model_overview}. We applied OLED to several benchmark datasets for anomaly detection in addition to providing a formal analysis of the efficacy of the Mask Module. Experimental results demonstrate that OLED is able to outperform a variety of recent state-of-the-art methods and hints at the broader usefulness of the mask module in other core computer vision tasks. In this paper our contributions are the following:
\begin{itemize}
    \item We proposed a novel approach for finding the most important parts of images for novelty detection.
    \item Our framework is optimized through adversarial setting which yields more efficient representations for novelty detection.
    \item Our method provides several anomaly scores which capture different aspects of normality
    \item Due to effectiveness of our method in masking important parts of the image, we can leverage it at the test time which yields better anomaly scores.
\end{itemize}


\section{Related Works}

One-class classification is primarily associated with the domain of novelty, outlier, and anomaly detection. In these types of problems, a model attempts to capture the distribution of the inlier class to finally detect the unknown outliers or novel concepts. The conventional methods in the anomaly detection field utilized one-class SVM \cite{scholkopf2002learning, hayton2000support} and Principal Component Analysis (PCA) and its variations \cite{bishop2006pattern, hoffmann2007kernel} to find a subspace that best represents the distribution of normal samples. Unsupervised clustering techniques like k-means \cite{zimek2012survey} and Gaussian Mixture Models (GMM) \cite{xiong2011group} also have been used to formulate the distribution of normal data for identifying the anomalies, but they normally fail in dealing with high-dimensional data.

Several other proposed methods benefit from self-representation learning, such as reconstruction-based approaches. They usually rely on the hypothesis that the outlier samples cannot be reconstructed precisely by a model that only learned the distribution of inlier samples. For example, Cong et al. \cite{cong2011sparse} suggested a model for video anomaly and outlier detection by learning sparse representations for distinguishing between inlier and outlier samples. In \cite{xu2015learning, sabokrou2016video}, test samples are reconstructed using the representations learned from inlier samples, and the reconstruction error is employed as a metric for novelty detection. Most of the deep learning-based models with encoder-decoder architecture \cite{sakurada2014anomaly, zhai2016deep, zhou2017anomaly, zong2018deep, chong2017abnormal} also used this score to detect anomalies. Although effective, these methods are limited by the under-designed representation of their latent space. Gong et al. \cite{gong2019memorizing} proposed a deep autoencoder augmented with a memory module to encode the input to a latent space with the encoder. The resulting latent vector is used as a query to retrieve the most relevant memory item for reconstruction with the decoder.

In \cite{schlegl2017unsupervised}, a deep convolutional generative adversarial network (GAN) is leveraged to learn a manifold of normal images with a novel anomaly score based on the mapping from image space to a random distribution. Sabokrou et al. \cite{sabokrou2018adversarially} took advantage of Generative Adversarial Networks (GAN) \cite{goodfellow2014generative} along with denoising autoencoders to use the discriminator’s score for the reconstructed images for the novelty detection task. Zaheer et al. \cite{zaheer2020old} redefined the adversarial one-class classifier training setup by modifying the role of the discriminator to distinguish between good and bad quality reconstructions and improved the results even further. Perera et al. used denoising auto-encoder networks to enforce the normal samples to be distributed uniformly across the latent space \cite{perera2019ocgan}. Abati et al. suggested a deep autoencoder model with a parametric density estimator that learns the probability distribution underlying its latent representations through an autoregressive procedure \cite{abati2019latent}.

Some recent works \cite{salehi2021multiresolution}, \cite{georgescu2021anomaly} have tried to leverage pre-trained deep neural networks by distilling the knowledge. In \cite{salehi2021multiresolution}, they utilized a VGG16 \cite{simonyan2014very} to compute a multi-level loss for training the student network to calculate the anomaly score and perform anomaly segmentation. Even though these methods could achieve high performance, they benefit from the knowledge attained by training on millions of labeled images and also may not work well on other modalities of data. As our proposed method does not leverage pretrained networks, we consider our work complimentary, and thus do not compare against this class of approaches.

\section{Method}

\subsection{Motivation}

Previous works have demonstrated that the reconstruction error of an Autoencoder (AE) acts as a good indicator of whether or not a sample conforms to the distribution defined in the training examples \cite{xia2015learning}. As such, AEs are commonly used for anomaly detection. To this end, Denoising Auotoencoders (DAE) have often been used because of the more robust representations they offer \cite{alain2014regularized}. Context Autoencoders (CAEs), a subclass of DAEs, have been particularly successful in the anomaly detection task by offering representations that capture the semantics of the underlying training distribution \cite{zimmerer2018context}.

However, CAEs have a number of disadvantages. The first drawback of CAEs is that they learn suboptimal representations by failing to consistently mask important parts of the image during training. Furthermore, they perform poorly at test time if they include random masking. This is because the mask placement is closely related to the reconstruction score. An outlier with a simple part of the image masked may have a lower reconstruction error than an inlier image with a difficult part of the image masked. Thus, random masking cannot be effectively used at test time for more robust anomaly detection. Conversely, our approach avoids these drawbacks by learning to mask intelligently. Experimental results from the ablation study in section 4.6 support this conclusion.  In order to mitigate these shortcomings while leveraging the benefits offered by CAEs, we propose a One-Class Learned Encoder-Decoder Network with adversarial context masking, which we call OLED.

\subsection{Overview}

 Our proposed framework, OLED, consists of two modules: the Reconstructor $R$ and the Mask Module $MM$. An overview of the architecture is available in \ref{model_overview}. $R$ and $MM$ are trained in an adversarial manner, where $R$ seeks to reconstruct images that have been covered by masks generated by $MM$. Masks have the same spatial resolution as input images with a single channel of 0 or 1 activations. As such, a masked image is easily obtained by taking an element-wise product of an image and its corresponding mask.

 Through the adversarial training process, $R$ learns representations that encode semantic information of the training distribution through the inpainting task. Alternatively, $MM$ learns to mask the most important parts of the input image by maximizing the reconstruction error of $R$. At test time, new samples are subjected to masks generated by $MM$ and fed to $R$ where the reconstruction error is used as an anomaly score. Accordingly, the reconstruction error will be low for the inlier class because $R$ is trained to reconstruct and inpaint inlier samples. However, in the case of anomalies, the reconstruction error will be higher primarily. This stems from the fact that $R$ learns to reconstruct and inpaint masked regions using mostly inlier samples.

\subsection{Reconstructor}

 $R$ is a convolutional encoder-decoder network that is trained to reconstruct masked images. Following some of the previous works  \cite{baur2020autoencoders}, a dense bottleneck is used. The full connectivity of the dense layer is helpful for the inpainting task, especially for shallow networks with low receptive fields. Moreover, $R$ does not include max-pooling layers for greater stability in training. To further promote stability, Leaky ReLU and batch normalization are used in each convolutional block. The values after the last convolution layer are clipped to in between -1 and 1.

\subsection{Mask Module}

$MM$ consists of a mask generator $MG$ followed by a threshold unit $T$ that generates masks of the same resolution as the input image. These masks are applied to the corresponding input image prior to being fed into $R$. $MM$ seeks to produce a mask that maximizes the reconstruction error of the input. In this way, it learns to mask the most optimal parts of the image. Thus, masks generated by $MM$ yield more comparable anomaly scores across samples in contrast to random masking.

\subsubsection{Mask Generator} $MG$ is a convolutional autoencoder that takes an input image and generates a corresponding activation map. This activation map is input into the threshold unit to produce a binary mask. Similar to $R$, $MG$ avoids the use of max pooling. Additionally, batch normalization and Leaky ReLU are used in each convolutional block, with the exception of the last convolution block that uses ReLU. In contrast to $R$, $MG$ has a spatial bottleneck and contains much fewer parameters. This reflects the fact that $R$ has a substantially more complex task than $MG$.

\subsubsection{Threshold Unit} Activation maps generated by $MG$ are input into $T$ to generate a mask. $T$ requires a threshold hyperparameter that determines what percentage of the pixels in the image will not be masked. In this way, the same amount of pixels are masked in each image, ensuring that the reconstruction errors are comparable between samples.

For each activation map, pixels with activations in the top 1 - $t$ percent are set to 0. The final mask is obtained by setting the remaining activations to one. More formally, given an activation map $A$ and a scalar $s$ that represents the numeric value of the pixel with the $t$ highest activations:

\begin{equation}
A_{ij} =
\begin{cases}
    0,  &\text{if } A_{ij}  \geq   s\\
    1,              & \text{otherwise}
\end{cases}
\end{equation}

As it stands, this is a discontinuous function, which is known to have less stable optimization. In order to eliminate this problem, the threshold operation is reformulated in terms of continuous ReLU activation function:

\begin{equation}
            A_{ij} =  \frac{max(A_{ij} * -1 + s, 0)}{max(A_{ij} * -1 + s, 0) + \epsilon}
\end{equation}

where max($A_{ij}$, 0) represents the ReLU activation, and $\epsilon$ is an infinitesimal positive scalar. The above formulation ensures continuity over the entire domain of the function enabling backpropagation through $T$ into $MG$.

\subsubsection{Masking Procedure}
$MG$ and $T$ sequentially process an input image to create a mask. Masks generated by $MM$ are single-channel binary images with the same spatial resolution as input images. The masked image is obtained by applying the mask to its corresponding image for each channel. More precisely, given an image $x$, the corresponding masked image $x_{m}$ is defined as:

\begin{equation}
            x_{m} =  x \odot MM(x)
\end{equation}

where $\odot$ denotes element-wise multiplication. In this way, activations in the mask that are 0 set the corresponding pixel in the input image to 0, otherwise the pixel remains unchanged. It is important to note that input images, and thus the reconstructions generated by $R$, are scaled between -1 and 1. Because of this, masked pixels are set to the midpoint of the color range.

\subsection{Adversarial Training}
Adversarial training is a learning mechanism in which two networks compete in a minmax game that iteratively enhances the ability to model the underlying distribution of the data. Following this intuition, Generative Adversarial Networks (GANs) \cite{goodfellow2014generative} have been proposed and shown immense success in generating samples with similar distribution of the training data. In order to do so, a generator network $G$ and discriminator network $D$ are trained in this manner. $G$ takes as input a noise vector and seeks to produce samples that follow the distribution of the training data. Alternatively, $D$ takes as input real samples from the training set along with fake samples generated by $G$ and seeks to discriminate between the two. More formally, given an image $x$ sampled from $p_{\text{data}}$ and a random latent vector $z$ sampled from $p_{z}$ the objective of a GAN is:
\begin{equation}
\begin{aligned}
\min_G \max_D
\mathbb{E}_{x \sim p_{\text{data}}(x)}[\log D(x)] + \\
\mathbb{E}_{z \sim p_{z}(z)}[\log (1 - D(G(z))]
\end{aligned}
\end{equation}

$G(z)$ is a sample generated by $G$ with input $z$. $D(x)$ and $D(G(z))$ are the discriminator's classification scores for a real and generated sample, respectively.

Similarly, we train $MM$ and $R$ adversarially. $MM$ seeks to generate masks that yield the highest reconstruction error from $R$. The total reconstruction error ${L}_{tot}$ consists of an L2 loss of the masked image ${L}_{mask}$, contextual loss of the masked region ${L}_{cont}$ and an L2 loss of an unperturbed image ${L}_{recon}$. Given an inlier image $x$ and the corresponding masked image ${x}_{m}$, ${L}_{mask}$, ${L}_{cont}$ and ${L}_{recon}$ are defined as:
\begin{equation}
{L}_{mask} = \left\Vert x - R({x}_{m}) \right\Vert^2
\end{equation}

\begin{equation}
{L}_{cont} = \left\Vert x_{c} - R({x}_{c}) \right\Vert
\end{equation}

\begin{equation}
{L}_{rec} = \left\Vert x - R(x) \right\Vert^2
\end{equation}

where $x_{c}$ is the masked region of the input image and $R({x}_{c})$ is the reconstruction of the masked region. $R({x}_{m})$ denotes the reconstruction of the masked image $x_{m}$. $R({x})$ is the reconstruction of the intact image $x$ with the Reconstructor. The following are the components of the objective:
\begin{itemize}
    \item ${L}_{mask}$: Forces the network to form a semantic understanding of characteristic elements of inlier samples.
    \item ${L}_{cont}$: Emphasizes that the masked region of the image is        reconstructed properly to avoid blurry reconstructions of the masked region.
    \item ${L}_{rec}$: Helps the network learn the distribution of unmasked inliers.
\end{itemize}

As such, the objective function of OLED is given by:
\begin{equation}
\begin{aligned}
\min_R \max_{MM}
L_{mask} + \gamma{L}_{cont} + \lambda {L}_{rec}
\end{aligned}
\end{equation}

where $\gamma$ and $\lambda$ are hyperparameters that weigh ${L}_{cont}$ and ${L}_{rec}$, respectively. Since $MM$ has no bearing on ${L}_{rec}$, it is not included in the error of $MM$.

\subsection{Anomaly Scoring}
 The three distinct loss terms in the OLED objective present the opportunity for three anomaly scores to be defined: ${s}_{mask}$, ${s}_{cont}$ and ${s}_{rec}$. ${s}_{mask}$, ${s}_{cont}$ and ${s}_{rec}$ are obtained through scaling ${L}_{mask}$, ${L}_{cont}$ and ${L}_{rec}$ between 0 and 1. By virtue of being derived from the respective losses, each anomaly score captures a different element of normality. $s_{cont}$ and $s_{mask}$ capture normality local to the masked region which tends to cover the most characteristic parts of the image. $s_{rec}$ captures the global normality of the image, taking into account how good the entire reconstruction of the image is. $s_{avg}$ is obtained by taking the average of ${s}_{mask}$, ${s}_{cont}$ and ${s}_{rec}$.

\section{Experiments}

This section contains a detailed analysis of the proposed method, OLED. In particular, we evaluated OLED on three datasets that are benchmarks in the novelty/anomaly detection literature, and the results are compared to recent state-of-the-art methods. Additionally, we presented a formal analysis exploring the effectiveness of masks generated by $MM$.

\subsection{Implementation Details}
OLED is implemented in Python using the TensorFlow  \cite{tensorflow2015-whitepaper}. A detailed overview of the architecture of $R$ and $MM$ is available in Section 3.3 and Section 3.4, respectively. $t$, $\lambda$ and $\gamma$ are set to 87.5, 1 and 50 respectively. These hyperparameters were set based on experimentation and an ablation study showing the stability of the performance across different settings. The threshold parameter can be adjusted based on the difficulty of the dataset; where larger values of the threshold are more suitable for more complex datasets. The weights of the the loss function listed as the defaults are to balance out the effect of reconstruction losses since they are on different scales. $R$ and $MM$ use an Adam optimizer with a learning rate of $5e^{-4}$, $b_{1}$ of .5 and  $b_{2}$ of .9. The networks are trained for 300 epochs. Following \cite{sabokrou2018adversarially},  a small validation set containing 150 samples from inliers and 150 samples from outliers from the training set is used to determine the best epoch to select models $R$ and $MM$.

\subsection{Datasets}
The three datasets chosen for the experiments are MNIST \cite{lecun1998mnist}, CIFAR-10 \cite{krizhevsky2009learning} and UCSD \cite{chan2008ucsd}. These particular datasets were chosen based on their popularity as benchmarks in the anomaly detection literature. The setups were chosen in a way that enables OLED to be compared to a variety of recent state-of-the-art methods.

{\bf MNIST:} MNIST is a dataset that contains 60,000 images of handwritten digits from 0 to 9. Images in MNIST are grayscale with a resolution of 28 x 28.

{\bf CIFAR-10:} CIFAR-10 is a dataset that contains 60,000 natural images of objects from across ten classes. Images in CIFAR-10 are RGB with a resolution of 32 x 32. Similar to MNIST, CIFAR-10 is also used widely as a benchmark in the anomaly detection literature. However, CIFAR-10 presents more of a challenge because images differ substantially across classes, and the background of images are not aligned.

{\bf UCSD:} This dataset \cite{li2013anomaly} consists of two subsets (Ped1 and Ped2) with different outdoor scenes. Available objects in the frames are pedestrians, cars, skateboarders, wheelchairs, and bicycles. Pedestrians are dominant in nearly all frames and considered as the normal class, while other objects are anomalies. We assessed our method on Ped2, which includes 2,550 frames in 16 training and 2,010 frames in 12 test videos, all with a resolution of 240×360 pixels. Following \cite{zaheer2020old}, we calculated frame-level area under the receiver operating characteristic  (AUCROC) and Equal Error Rate (EER) to evaluate performance and compare against both patch-based and full-frame setups.

\subsection{Novelty Detection in Image Datasets}

{\bf MNIST:} OLED is evaluated on MNIST using the protocol defined in \cite{gong2019memorizing}. This protocol involves dividing the dataset into ten different anomaly detection datasets corresponding to the ten predefined classes in MNIST. In each anomaly detection dataset, the inliers are sampled from 1 class, and the outliers are sampled from the remaining 9 classes. The normal data is split into train and test sets with a ratio of 2:1, and the anomaly proposition is set to be 30\%. Following \cite{gong2019memorizing}, AUCROC is the evaluation metric for this experiment.

\begin{table}
\begin{center}
 \begin{tabular}{|l|c|}
 \hline
Method & AUCROC \\ [0.5ex]
 \hline\hline
 OCSVM \cite{scholkopf2002learning} & 0.9499  \\

 AE \cite{bengio2007greedy} & 0.9619 \\

 VAE \cite{kingma2013auto} & 0.9643  \\

 PixCNN  \cite{oord2016conditional} & 0.6141  \\

 DSEBM \cite{zhai2016deep} & 0.9554 \\

 MemAE \cite{gong2019memorizing} & 0.9751 \\

 \hline
 OLED (Ours) $s_{rec}$  & \textbf{0.9772}  \\
 OLED (Ours) $s_{mask}$  & \textbf{0.9851} \\
 OLED (Ours) $s_{cont}$  & 0.9650 \\
 OLED (Ours) $s_{avg}$  &  \textbf{0.9845} \\
 \hline
\end{tabular}
\end{center}
\caption{Average AUCROC values on all 10 classes sampled from MNIST image dataset.}
\label{mnist_exp}
\end{table}

\begin{figure}[t]
\begin{center}
   \includegraphics[width=1\linewidth]{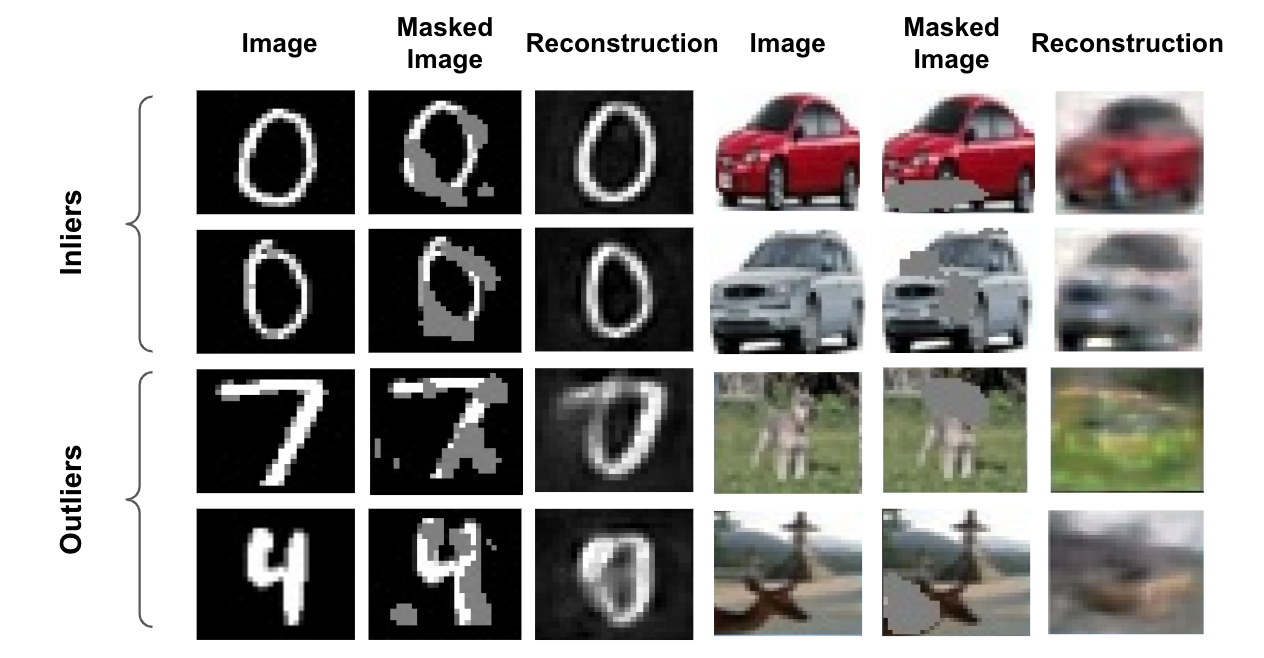}
\end{center}
   \caption{OLED Reconstructions. For both MNIST and CIFAR-10, the original image, perturbed image after applying the mask generated by $MM$ (masks are illustrated in gray) and the final reconstruction are shown. Inlier samples are in the top two rows and outlier samples are in the bottom two rows.}
\label{CIFAR-10_mnist}
\end{figure}

Given the protocol in \cite{gong2019memorizing}, OLED is compared against MemAE \cite{gong2019memorizing} and other methods \cite{scholkopf2002learning, kingma2013auto, oord2016conditional, zhai2016deep}. The results are reported in Table \ref{mnist_exp}. OLED yields excellent results, surpassing MemAE and other approaches. In particular,  $s_{rec}$, $s_{mask}$ and $s_{avg}$ exceed all other identified approaches, recording an AUCROC of 0.977, 0.985 and 0.984, respectively. A visualization of OLED applied to both inlier and outlier samples for MNIST is available in Figure \ref{CIFAR-10_mnist}. Additionally, in Figure \ref{oled_ae}, the reconstructions of OLED are compared to that of a normal AE, further demonstrating the superiority of the representations offered by OLED for the anomaly detection task.

{\bf CIFAR-10:} OLED is evaluated on CIFAR-10 using the protocol defined in \cite{perera2019ocgan}. This protocol involves dividing the dataset into ten different anomaly detection datasets corresponding to the ten predefined classes in CIFAR-10. In each anomaly detection dataset, the inliers are sampled from 1 class, and the outliers are sampled from the remaining 9 classes. The predefined train and test splits are used to conduct the experiments. Testing data of all classes are used for testing. Following \cite{perera2019ocgan}, AUCROC is the evaluation metric for this experiment.

\begin{table}
\begin{center}
 \begin{tabular}{|l|c|}
 \hline
Method & AUCROC \\ [0.5ex]
 \hline\hline
 OCSVM \cite{scholkopf2002learning} & 0.5856  \\

 DAE \cite{vincent2010stacked} & 0.5358  \\

 VAE \cite{kingma2013auto} & 0.5833  \\

 PixCNN \cite{oord2016conditional} & 0.5506  \\

 GAN \cite{schlegl2017unsupervised} & 0.5916 \\

 AND \cite{abati2019latent} & 0.6172  \\

 AnoGAN \cite{schlegl2017unsupervised} & 0.6179 \\

 DSVDD \cite{ruff2018deep} & 0.6481  \\

 OCGAN \cite{perera2019ocgan} & 0.6566  \\

 \hline
 OLED (Ours) $s_{rec}$  & \textbf{0.6622} \\
 OLED (Ours) $s_{mask}$  & \textbf{0.6711} \\
 OLED (Ours) $s_{avg}$  & \textbf{0.6683} \\
 OLED (Ours) $s_{cont}$  & \textbf{0.6673} \\

 \hline
\end{tabular}
\end{center}
\caption{One-class novelty detection Average AUCROC results on CIFAR-10 image dataset following the protocol in \cite{perera2019ocgan}.}
\label{CIFAR-10_exp}
\end{table}

OLED is compared to OCGAN \cite{perera2019ocgan} and other recently proposed methods for anomaly detection \cite{abati2019latent, vincent2010stacked, schlegl2017unsupervised, ruff2018deep}. The results are reported in Table \ref{CIFAR-10_exp}. OLED outperforms the compared methods, including OCGAN, by a considerable margin. Particularly, $s_{rec}$, $s_{mask}$, $s_{avg}$ and $s_{cont}$ exceed all other identified approaches, recording an AUCROC of 0.662, 0.671, 0.6683 and  0.667, respectively. A visualization of OLED applied to both inlier and outlier samples for CIFAR-10 is available in Figure \ref{CIFAR-10_mnist}.

\subsection{Video Novelty Detection}
\begin{figure}[t]
\begin{center}
   \includegraphics[width=1\linewidth]{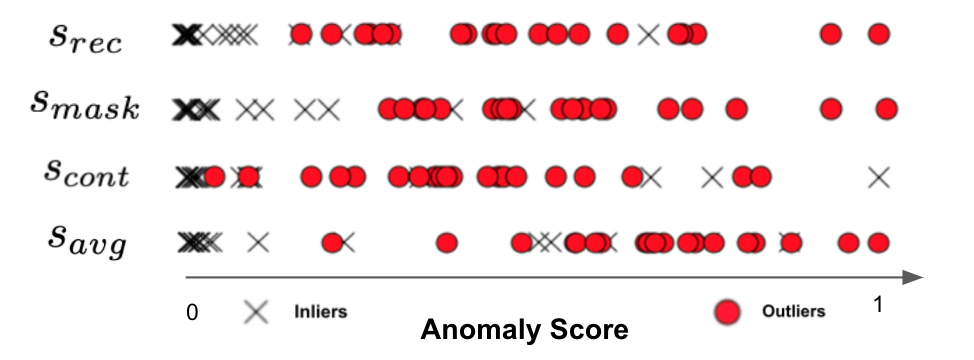}
\end{center}
\caption{Sample of anomaly scores for both the inlier and outlier class for the UCSD dataset.}
\label{ucsd_scores}
\end{figure}

One of the common use cases of one-class classification is in the domain of novelty detection for surveillance purposes \cite{gong2019memorizing,zhou2017anomaly, sabokrou2018adversarially}. Nonetheless, this task is more difficult in the video domain because of the variations of mobile objects across the frames. In this experiment, each frame of the dataset is divided into patches of size 30×30 pixels following \cite{sabokrou2018adversarially}. Training patches only include scenes of walking pedestrians, while in the testing phase, patches are extracted from outlier frames that contain abnormal as well as normal objects. Frame-level  AUROC and EER are the two metrics used to compare our method with state-of-the-art methods in recent years. As depicted in Table \ref{ucsd_exp}, our method outperforms recent state-of-the-art models in the video novelty detection task. More specifically, our approach achieves an AUCROC performance of 99.02\% and an EER of 5.4\%. The visualization in Figure \ref{ucsd_scores} demonstrates the separability of anomaly scores for inliers and outliers.

\begin{table}
\begin{center}
 \begin{tabular}{|l|l|c|}
 \hline
Method & AUCROC & EER \\ [0.5ex]
 \hline\hline
TSC \cite{luo2017revisit_novelty}                       & 0.922 & -                  \\
FRCN action \cite{hinami2017joint_novelty}               & 0.922                  & -                  \\
AbnormalGAN \cite{ravanbakhsh2017abnormal_novelty}               & 0.935                   & 0.13               \\
MemAE \cite{gong2019memorizing} & 0.941                   & -                \\
GrowingGas \cite{sun2017online}                & 0.941   & -                \\
FFP \cite{liu2018future_novelty}     & 0.954  & -                  \\
ConvAE+UNet \cite{Nguyen_2019_ICCV}               & 0.962  & -                  \\
STAN \cite{lee2018stan}        & 0.965   &   -    \\
Object-centric \cite{ionescu2019object} &  0.978   & -                       \\
Ravanbakhsh \cite{ravanbakhsh2019training} & - & 0.14       \\
ALOCC \cite{sabokrou2018adversarially} & - & 0.13       \\
Deep-cascade \cite{sabokrou2017deep_novelty} & - & 0.09       \\
 Old is gold \cite{zaheer2020old} & 0.981 & 0.07       \\

 \hline
 OLED (Ours) $s_{rec}$  & \textbf{0.9854} & \textbf{0.0646} \\
 OLED (Ours) $s_{mask}$  & \textbf{0.9853} & \textbf{0.0683} \\
 OLED (Ours) $s_{avg}$  & \textbf{0.9902} & \textbf{0.0540} \\
 OLED (Ours) $s_{cont}$  & \textbf{0.9866} & \textbf{0.0606} \\

 \hline
\end{tabular}
\end{center}
\caption{Frame-level AUCROC and EER comparison on UCSD dataset with state-of-the-art methods.}
\label{ucsd_exp}
\end{table}

\subsection{Mask Module Evaluation}
\begin{figure}[t]
\begin{center}
   \includegraphics[width=1\linewidth]{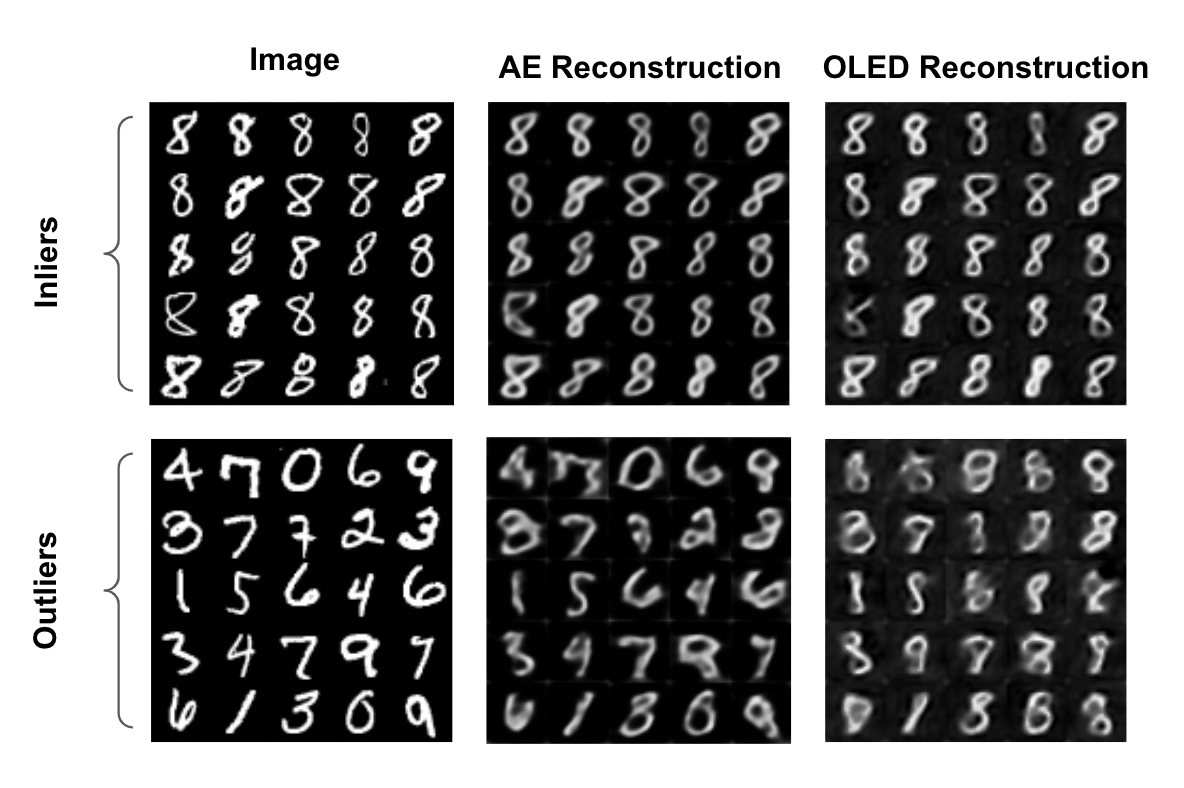}
\end{center}
\caption{AE vs OLED Reconstructions for the MNIST dataset.}
\label{oled_ae}
\end{figure}

The results from the experiments in Section 4.3 and Section 4.4 are a clear indication that $OLED$ is a strong method for anomaly detection. In every case, anomaly scores that leveraged masking, and by extension $MM$, yielded the highest performance. Visual results in Figure \ref{CIFAR-10_mnist} and \ref{oled_ae} support the initial hypothesis that $MM$ generates masks that cover important structures in the input image. Furthermore, this is the case for both inlier and outlier images. The following section seeks to solidify these observations more formally.

To quantitatively assess the effectiveness of $MM$ in masking important parts of images, $MM$ is re-purposed to perform a binary segmentation task that involves identifying whether or not each pixel in the input image is important. Specifically, the activation maps $A$ generated by $MM$ serve as the predicted semantic maps for images. $A$ is used instead of $MM(x)$ to avoid the threshold constraint imposed by $T$. Using $A$ and the ground truth semantic maps, the pixelwise AUCROC score is computed for both inlier and outlier images.

The aforementioned analysis is realized by evaluating the $MM$ trained on digit class 8 from the MNIST experiments in Section 4.3 on the corresponding test set. MNIST is well suited for this experiment because we are able to make the assumption that nonzero pixels are part of the digit and thus important. The ground truth semantic maps for the test set are obtained by setting non zero activations to 1 otherwise 0.
The former signals the pixel corresponds to part of the written digit, and the latter signals the pixel is part of the background.

The results for the experiment are displayed in Table \ref{segment_exp}. $MM$ is able to segment important pixels in both inlier and outlier images with a high degree of accuracy with no modifications to the original architecture. This is a testament to the usefulness of $MM$ in the anomaly detection task and hints at broader use cases in computer vision.

\subsection{Ablation Study}
In order to further assess the value of the proposed learned masking approach, OLED is compared to the baseline method context autoencoders (CAE). As CAEs employ random masking during training, the following section seeks to compare the learned masking proposed by OLED with random masking utilized in CAEs. To realize this comparison, a CAE was implemented and evaluated on the MNIST dataset using the protocol outlined in Section 4.3. The CAE shared the same architecture as $R$. The CAE is given input images with a random 10 x 10 region cropped out during training, keeping the number of masked pixels relatively consistent with $R$.

The results from the above experiment are displayed in \ref{oled_vs_cae}. Similar to OLED, $s_{rec}$, $s_{mask}$, $s_{avg}$ and $s_{cont}$ are reported for the CAE. OLED is able to substantially outperform CAE, despite having identical architectures for the base reconstruction module. This is a clear indication that the learned masking approach proposed in OLED outperforms random masking for the anomaly detection task. Additionally, masking at test time enhances the performance of OLED but substantially decreases the performance of the CAE. This supports our intuition that the wrong placement of the masks by CAEs leads to suboptimal representations and introduce unwanted variations in the reconstruction error of samples that is detrimental to novelty detection performance.

\begin{table}
\begin{center}
 \begin{tabular}{|l|l|c|}
 \hline
Method & Score Type & AUCROC \\ [0.5ex]
 \hline\hline
 CAE & $s_{rec}$  & 0.9209  \\
 CAE & $s_{mask}$  & 0.8936 \\
 CAE & $s_{cont}$  & 0.6869 \\
 CAE & $s_{avg}$  &  0.8768 \\
 \hline
 OLED (Ours) & $s_{rec}$  & \textbf{0.9772}  \\
 OLED (Ours) & $s_{mask}$  & \textbf{0.9851} \\
 OLED (Ours) & $s_{cont}$  & \textbf{0.9650} \\
 OLED (Ours) & $s_{avg}$  &  \textbf{0.9845} \\
 \hline
\end{tabular}
\end{center}
\caption{Comparison between our method (OLED) vs. Context Autoencoder (CAE) on MNIST image dataset.}
\label{oled_vs_cae}
\end{table}

\begin{table}
\begin{center}
 \begin{tabular}{|l|c|}
 \hline
 Data  & AUCROC \\ [0.5ex]
 \hline\hline
 Inlier & 0.8499  \\
 Outlier & 0.8472 \\
 \hline
\end{tabular}
\end{center}
\caption{Segmentaion performance of mask generator $M$ on MNIST dataset.}
\label{segment_exp}
\end{table}

\section{Discussion}
The results presented in Section 4 are a clear indication of the effectiveness of OLED for the anomaly detection task. In all three anomaly detection experiments on MNIST, CIFAR-10 and UCSD, OLED outperformed state-of-the-art methods by a large margin. Additional experiments evaluating the performance of $MM$ demonstrated strong performance in segmenting the most important parts of samples for both the inlier and outlier class.

As initially hypothesized, OLED is able to reconstruct samples from the inlier class with ease but struggles to reconstruct samples from the outlier class. This addresses one of the fundamental problems AE face when applied to the anomaly detection task; reconstructing outliers too well. OLED accomplishes this by offering representations that are optimized for reconstructing important parts of the inlier samples through the adversarial training of $R$ and $MM$. Beyond this, anomaly detection is enhanced through the use of masking at test time.

OLED also presents the benefit of being trained end-to-end, resulting in a less cumbersome training procedure than some of the identified methods. In this way, $MM$ can be included seamlessly into existing AE-based anomaly detection methods. There are also no constraints that prevent OLED from being applied to other modalities of data.  Furthermore, the core innovation proposed in this research, learned optimal masking, has the potential to be applied to other tasks in computer vision and beyond.

\section{Conclusion}
In this paper, we proposed an adversarial framework for novelty detection in both images and videos. More specifically, our method includes a Mask Module and a Reconstructor; the Mask Module is a convolutional autoencoder that learns to cover the most important parts of images, and the Reconstructor is a convolutional encoder-decoder that strives to reconstruct the masked images. The mask module will learn to mask the parts of input in a way to increase the reconstruction loss while the Reconstructor tries to minimize it. The proposed approach allows semantically rich representations and improves novelty detection at test time by covering the most important parts of the context. We have applied our method to a variety of tasks, including outlier and anomaly detection in images and videos. The results illustrate the superiority of OLED in identifying samples related to the outlier class compared to recent state-of-the-art models.


{\small
\bibliography{egbib}

\begin{thebibliography}{10}

\bibitem{schlegl2017unsupervised}
Thomas Schlegl, Philipp Seeb{\"o}ck, Sebastian~M Waldstein, Ursula
  Schmidt-Erfurth, and Georg Langs.
\newblock Unsupervised anomaly detection with generative adversarial networks
  to guide marker discovery.
\newblock In {\em International conference on information processing in medical
  imaging}, pages 146--157. Springer, 2017.

\bibitem{luo2017revisit}
Weixin Luo, Wen Liu, and Shenghua Gao.
\newblock A revisit of sparse coding based anomaly detection in stacked rnn
  framework.
\newblock In {\em Proceedings of the IEEE International Conference on Computer
  Vision}, pages 341--349, 2017.

\bibitem{zimek2012survey}
Arthur Zimek, Erich Schubert, and Hans-Peter Kriegel.
\newblock A survey on unsupervised outlier detection in high-dimensional
  numerical data.
\newblock {\em Statistical Analysis and Data Mining: The ASA Data Science
  Journal}, 5(5):363--387, 2012.

\bibitem{bengio2013representation}
Yoshua Bengio, Aaron Courville, and Pascal Vincent.
\newblock Representation learning: A review and new perspectives.
\newblock {\em IEEE transactions on pattern analysis and machine intelligence},
  35(8):1798--1828, 2013.

\bibitem{bengio2007greedy}
Yoshua Bengio, Pascal Lamblin, Dan Popovici, Hugo Larochelle, et~al.
\newblock Greedy layer-wise training of deep networks.
\newblock {\em Advances in neural information processing systems}, 19:153,
  2007.

\bibitem{chalapathy2019deep}
Raghavendra Chalapathy and Sanjay Chawla.
\newblock Deep learning for anomaly detection: A survey.
\newblock {\em arXiv preprint arXiv:1901.03407}, 2019.

\bibitem{xia2015learning}
Yan Xia, Xudong Cao, Fang Wen, Gang Hua, and Jian Sun.
\newblock Learning discriminative reconstructions for unsupervised outlier
  removal.
\newblock In {\em Proceedings of the IEEE International Conference on Computer
  Vision}, pages 1511--1519, 2015.

\bibitem{zong2018deep}
Bo~Zong, Qi~Song, Martin~Renqiang Min, Wei Cheng, Cristian Lumezanu, Daeki Cho,
  and Haifeng Chen.
\newblock Deep autoencoding gaussian mixture model for unsupervised anomaly
  detection.
\newblock In {\em International Conference on Learning Representations}, 2018.

\bibitem{gong2019memorizing}
Dong Gong, Lingqiao Liu, Vuong Le, Budhaditya Saha, Moussa~Reda Mansour, Svetha
  Venkatesh, and Anton van~den Hengel.
\newblock Memorizing normality to detect anomaly: Memory-augmented deep
  autoencoder for unsupervised anomaly detection.
\newblock In {\em Proceedings of the IEEE/CVF International Conference on
  Computer Vision}, pages 1705--1714, 2019.

\bibitem{abati2019latent}
Davide Abati, Angelo Porrello, Simone Calderara, and Rita Cucchiara.
\newblock Latent space autoregression for novelty detection.
\newblock In {\em Proceedings of the IEEE/CVF Conference on Computer Vision and
  Pattern Recognition}, pages 481--490, 2019.

\bibitem{perera2019ocgan}
Pramuditha Perera, Ramesh Nallapati, and Bing Xiang.
\newblock Ocgan: One-class novelty detection using gans with constrained latent
  representations.
\newblock In {\em Proceedings of the IEEE/CVF Conference on Computer Vision and
  Pattern Recognition}, pages 2898--2906, 2019.

\bibitem{zimmerer2018context}
David Zimmerer, Simon~AA Kohl, Jens Petersen, Fabian Isensee, and Klaus~H
  Maier-Hein.
\newblock Context-encoding variational autoencoder for unsupervised anomaly
  detection.
\newblock {\em arXiv preprint arXiv:1812.05941}, 2018.

\bibitem{vincent2010stacked}
Pascal Vincent, Hugo Larochelle, Isabelle Lajoie, Yoshua Bengio, Pierre-Antoine
  Manzagol, and L{\'e}on Bottou.
\newblock Stacked denoising autoencoders: Learning useful representations in a
  deep network with a local denoising criterion.
\newblock {\em Journal of machine learning research}, 11(12), 2010.

\bibitem{alain2014regularized}
Guillaume Alain and Yoshua Bengio.
\newblock What regularized auto-encoders learn from the data-generating
  distribution.
\newblock {\em The Journal of Machine Learning Research}, 15(1):3563--3593,
  2014.

\bibitem{pathak2016context}
Deepak Pathak, Philipp Krahenbuhl, Jeff Donahue, Trevor Darrell, and Alexei~A
  Efros.
\newblock Context encoders: Feature learning by inpainting.
\newblock In {\em Proceedings of the IEEE conference on computer vision and
  pattern recognition}, pages 2536--2544, 2016.

\bibitem{baur2020autoencoders}
Christoph Baur, Stefan Denner, Benedikt Wiestler, Nassir Navab, and Shadi
  Albarqouni.
\newblock Autoencoders for unsupervised anomaly segmentation in brain mr
  images: A comparative study.
\newblock {\em Medical Image Analysis}, page 101952, 2020.

\bibitem{scholkopf2002learning}
Bernhard Sch{\"o}lkopf, Alexander~J Smola, Francis Bach, et~al.
\newblock {\em Learning with kernels: support vector machines, regularization,
  optimization, and beyond}.
\newblock MIT press, 2002.

\bibitem{hayton2000support}
Paul Hayton, Bernhard Sch{\"o}lkopf, Lionel Tarassenko, and Paul Anuzis.
\newblock Support vector novelty detection applied to jet engine vibration
  spectra.
\newblock In {\em NIPS}, pages 946--952. Citeseer, 2000.

\bibitem{bishop2006pattern}
Christopher~M Bishop.
\newblock {\em Pattern recognition and machine learning}.
\newblock springer, 2006.

\bibitem{hoffmann2007kernel}
Heiko Hoffmann.
\newblock Kernel pca for novelty detection.
\newblock {\em Pattern recognition}, 40(3):863--874, 2007.

\bibitem{xiong2011group}
Liang Xiong, Barnab{\'a}s P{\'o}czos, and Jeff Schneider.
\newblock Group anomaly detection using flexible genre models.
\newblock In {\em Proceedings of the 24th International Conference on Neural
  Information Processing Systems}, 2011.

\bibitem{cong2011sparse}
Yang Cong, Junsong Yuan, and Ji~Liu.
\newblock Sparse reconstruction cost for abnormal event detection.
\newblock In {\em CVPR 2011}, pages 3449--3456. IEEE, 2011.

\bibitem{xu2015learning}
Dan Xu, Elisa Ricci, Yan Yan, Jingkuan Song, and Nicu Sebe.
\newblock Learning deep representations of appearance and motion for anomalous
  event detection.
\newblock {\em arXiv preprint arXiv:1510.01553}, 2015.

\bibitem{sabokrou2016video}
Mohammad Sabokrou, Mahmood Fathy, and Mojtaba Hoseini.
\newblock Video anomaly detection and localisation based on the sparsity and
  reconstruction error of auto-encoder.
\newblock {\em Electronics Letters}, 52(13):1122--1124, 2016.

\bibitem{sakurada2014anomaly}
Mayu Sakurada and Takehisa Yairi.
\newblock Anomaly detection using autoencoders with nonlinear dimensionality
  reduction.
\newblock In {\em Proceedings of the MLSDA 2014 2nd Workshop on Machine
  Learning for Sensory Data Analysis}, pages 4--11, 2014.

\bibitem{zhai2016deep}
Shuangfei Zhai, Yu~Cheng, Weining Lu, and Zhongfei Zhang.
\newblock Deep structured energy based models for anomaly detection.
\newblock In {\em International Conference on Machine Learning}, pages
  1100--1109. PMLR, 2016.

\bibitem{zhou2017anomaly}
Chong Zhou and Randy~C Paffenroth.
\newblock Anomaly detection with robust deep autoencoders.
\newblock In {\em Proceedings of the 23rd ACM SIGKDD international conference
  on knowledge discovery and data mining}, pages 665--674, 2017.

\bibitem{chong2017abnormal}
Yong~Shean Chong and Yong~Haur Tay.
\newblock Abnormal event detection in videos using spatiotemporal autoencoder.
\newblock In {\em International symposium on neural networks}, pages 189--196.
  Springer, 2017.

\bibitem{sabokrou2018adversarially}
Mohammad Sabokrou, Mohammad Khalooei, Mahmood Fathy, and Ehsan Adeli.
\newblock Adversarially learned one-class classifier for novelty detection.
\newblock In {\em Proceedings of the IEEE Conference on Computer Vision and
  Pattern Recognition}, pages 3379--3388, 2018.

\bibitem{goodfellow2014generative}
Ian~J Goodfellow, Jean Pouget-Abadie, Mehdi Mirza, Bing Xu, David Warde-Farley,
  Sherjil Ozair, Aaron~C Courville, and Yoshua Bengio.
\newblock Generative adversarial nets.
\newblock In {\em NIPS}, 2014.

\bibitem{zaheer2020old}
Muhammad~Zaigham Zaheer, Jin-ha Lee, Marcella Astrid, and Seung-Ik Lee.
\newblock Old is gold: Redefining the adversarially learned one-class
  classifier training paradigm.
\newblock In {\em Proceedings of the IEEE/CVF Conference on Computer Vision and
  Pattern Recognition}, pages 14183--14193, 2020.

\bibitem{salehi2021multiresolution}
Mohammadreza Salehi, Niousha Sadjadi, Soroosh Baselizadeh, Mohammad~H Rohban,
  and Hamid~R Rabiee.
\newblock Multiresolution knowledge distillation for anomaly detection.
\newblock In {\em Proceedings of the IEEE/CVF Conference on Computer Vision and
  Pattern Recognition}, pages 14902--14912, 2021.

\bibitem{georgescu2021anomaly}
Mariana-Iuliana Georgescu, Antonio Barbalau, Radu~Tudor Ionescu, Fahad~Shahbaz
  Khan, Marius Popescu, and Mubarak Shah.
\newblock Anomaly detection in video via self-supervised and multi-task
  learning.
\newblock In {\em Proceedings of the IEEE/CVF Conference on Computer Vision and
  Pattern Recognition}, pages 12742--12752, 2021.

\bibitem{simonyan2014very}
Karen Simonyan and Andrew Zisserman.
\newblock Very deep convolutional networks for large-scale image recognition.
\newblock {\em arXiv preprint arXiv:1409.1556}, 2014.

\bibitem{tensorflow2015-whitepaper}
Mart\'{\i}n Abadi, Ashish Agarwal, Paul Barham, Eugene Brevdo, Zhifeng Chen,
  Craig Citro, Greg~S. Corrado, Andy Davis, Jeffrey Dean, Matthieu Devin,
  Sanjay Ghemawat, Ian Goodfellow, Andrew Harp, Geoffrey Irving, Michael Isard,
  Yangqing Jia, Rafal Jozefowicz, Lukasz Kaiser, Manjunath Kudlur, Josh
  Levenberg, Dandelion Man\'{e}, Rajat Monga, Sherry Moore, Derek Murray, Chris
  Olah, Mike Schuster, Jonathon Shlens, Benoit Steiner, Ilya Sutskever, Kunal
  Talwar, Paul Tucker, Vincent Vanhoucke, Vijay Vasudevan, Fernanda Vi\'{e}gas,
  Oriol Vinyals, Pete Warden, Martin Wattenberg, Martin Wicke, Yuan Yu, and
  Xiaoqiang Zheng.
\newblock {TensorFlow}: Large-scale machine learning on heterogeneous systems,
  2015.
\newblock Software available from tensorflow.org.

\bibitem{lecun1998mnist}
Yann LeCun.
\newblock The mnist database of handwritten digits.
\newblock {\em http://yann. lecun. com/exdb/mnist/}, 1998.

\bibitem{krizhevsky2009learning}
Alex Krizhevsky, Geoffrey Hinton, et~al.
\newblock Learning multiple layers of features from tiny images.
\newblock {\em Technical Report}, 2009.

\bibitem{chan2008ucsd}
Antoni Chan and Nuno Vasconcelos.
\newblock Ucsd pedestrian dataset.
\newblock {\em IEEE Trans. on Pattern Analysis and Machine Intelligence
  (TPAMI)}, 30(5):909--926, 2008.

\bibitem{li2013anomaly}
Weixin Li, Vijay Mahadevan, and Nuno Vasconcelos.
\newblock Anomaly detection and localization in crowded scenes.
\newblock {\em IEEE transactions on pattern analysis and machine intelligence},
  36(1):18--32, 2013.

\bibitem{kingma2013auto}
Diederik~P Kingma and Max Welling.
\newblock Auto-encoding variational bayes.
\newblock {\em arXiv preprint arXiv:1312.6114}, 2013.

\bibitem{oord2016conditional}
Aaron van~den Oord, Nal Kalchbrenner, Oriol Vinyals, Lasse Espeholt, Alex
  Graves, and Koray Kavukcuoglu.
\newblock Conditional image generation with pixelcnn decoders.
\newblock {\em arXiv preprint arXiv:1606.05328}, 2016.

\bibitem{ruff2018deep}
Lukas Ruff, Robert Vandermeulen, Nico Goernitz, Lucas Deecke, Shoaib~Ahmed
  Siddiqui, Alexander Binder, Emmanuel M{\"u}ller, and Marius Kloft.
\newblock Deep one-class classification.
\newblock In {\em International conference on machine learning}, pages
  4393--4402. PMLR, 2018.

\bibitem{luo2017revisit_novelty}
Weixin Luo, Wen Liu, and Shenghua Gao.
\newblock A revisit of sparse coding based anomaly detection in stacked rnn
  framework.
\newblock In {\em Proceedings of the IEEE International Conference on Computer
  Vision}, pages 341--349, 2017.

\bibitem{hinami2017joint_novelty}
Ryota Hinami, Tao Mei, and Shin'ichi Satoh.
\newblock Joint detection and recounting of abnormal events by learning deep
  generic knowledge.
\newblock In {\em Proceedings of the IEEE International Conference on Computer
  Vision}, pages 3619--3627, 2017.

\bibitem{ravanbakhsh2017abnormal_novelty}
Mahdyar Ravanbakhsh, Moin Nabi, Enver Sangineto, Lucio Marcenaro, Carlo
  Regazzoni, and Nicu Sebe.
\newblock Abnormal event detection in videos using generative adversarial nets.
\newblock In {\em 2017 IEEE International Conference on Image Processing
  (ICIP)}, pages 1577--1581. IEEE, 2017.

\bibitem{sun2017online}
Qianru Sun, Hong Liu, and Tatsuya Harada.
\newblock Online growing neural gas for anomaly detection in changing
  surveillance scenes.
\newblock {\em Pattern Recognition}, 64:187--201, 2017.

\bibitem{liu2018future_novelty}
Wen Liu, Weixin Luo, Dongze Lian, and Shenghua Gao.
\newblock Future frame prediction for anomaly detection--a new baseline.
\newblock In {\em Proceedings of the IEEE Conference on Computer Vision and
  Pattern Recognition}, pages 6536--6545, 2018.

\bibitem{Nguyen_2019_ICCV}
Trong-Nguyen Nguyen and Jean Meunier.
\newblock Anomaly detection in video sequence with appearance-motion
  correspondence.
\newblock In {\em The IEEE International Conference on Computer Vision (ICCV)},
  October 2019.

\bibitem{lee2018stan}
Sangmin Lee, Hak~Gu Kim, and Yong~Man Ro.
\newblock Stan: Spatio-temporal adversarial networks for abnormal event
  detection.
\newblock In {\em 2018 IEEE International Conference on Acoustics, Speech and
  Signal Processing (ICASSP)}, pages 1323--1327. IEEE, 2018.

\bibitem{ionescu2019object}
Radu~Tudor Ionescu, Fahad~Shahbaz Khan, Mariana-Iuliana Georgescu, and Ling
  Shao.
\newblock Object-centric auto-encoders and dummy anomalies for abnormal event
  detection in video.
\newblock In {\em Proceedings of the IEEE Conference on Computer Vision and
  Pattern Recognition}, pages 7842--7851, 2019.

\bibitem{ravanbakhsh2019training}
Mahdyar Ravanbakhsh, Enver Sangineto, Moin Nabi, and Nicu Sebe.
\newblock Training adversarial discriminators for cross-channel abnormal event
  detection in crowds.
\newblock In {\em 2019 IEEE Winter Conference on Applications of Computer
  Vision (WACV)}, pages 1896--1904. IEEE, 2019.

\bibitem{sabokrou2017deep_novelty}
Mohammad Sabokrou, Mohsen Fayyaz, Mahmood Fathy, and Reinhard Klette.
\newblock Deep-cascade: Cascading 3d deep neural networks for fast anomaly
  detection and localization in crowded scenes.
\newblock {\em IEEE Transactions on Image Processing}, 26(4):1992--2004, 2017.

\end{thebibliography}
}

\end{document}